\documentclass[10pt,twocolumn,letterpaper]{article}

\usepackage{cvpr}              
\usepackage{times}
\usepackage{epsfig}
\usepackage{graphicx}
\usepackage{amsmath}
\usepackage{amssymb}
\usepackage{capt-of}
\usepackage{multirow}
\usepackage{amsthm}
\usepackage{comment}
\usepackage{diagbox}
\usepackage{wasysym}
\usepackage{pifont}
\usepackage{bbm, dsfont}
\usepackage{caption}
\usepackage{float}

\usepackage{color}
\usepackage{xcolor}
\definecolor{cvprblue}{rgb}{0.21,0.49,0.74}
\usepackage[pagebackref,breaklinks,colorlinks,citecolor=cvprblue]{hyperref}

\usepackage[capitalize]{cleveref}
\crefname{section}{Sec.}{Secs.}
\Crefname{section}{Section}{Sections}
\Crefname{table}{Table}{Tables}
\crefname{table}{Tab.}{Tabs.}

\def\paperID{} 
\def\confName{3DV\xspace}
\def\confYear{2024\xspace}

\newlength\savewidth\newcommand\shline{\noalign{\global\savewidth\arrayrulewidth
  \global\arrayrulewidth 1pt}\hline\noalign{\global\arrayrulewidth\savewidth}}
\newcommand{\tablestyle}[2]{\setlength{\tabcolsep}{#1}\renewcommand{\arraystretch}{#2}\centering\footnotesize}

\newcommand\blfootnote[1]{%
  \begingroup
  \renewcommand\thefootnote{}\footnote{#1}%
  \addtocounter{footnote}{-1}%
  \endgroup
}

\begin{document}
\title{ContactArt: Learning 3D Interaction Priors for \\ Category-level Articulated Object and Hand Poses Estimation}

\author{
Zehao Zhu$^{1*}$ \quad 
Jiashun Wang$^{2*}$ \quad 
Yuzhe Qin$^{3}$ \quad 
Deqing Sun$^{4}$ \quad 
Varun Jampani$^{4}$ \quad 
Xiaolong Wang$^{3}$\\
$^1$ University of Texas at Austin~~~
$^2$Carnegie Mellon University~~~
$^3$UC San Diego~~~ 
$^4$Google Research~~~
\\}

\twocolumn[{
\maketitle
\renewcommand\twocolumn[1][]{#1}
\vspace{-2.5em}
\begin{center}
    \centering
    \includegraphics[width=0.92\textwidth]{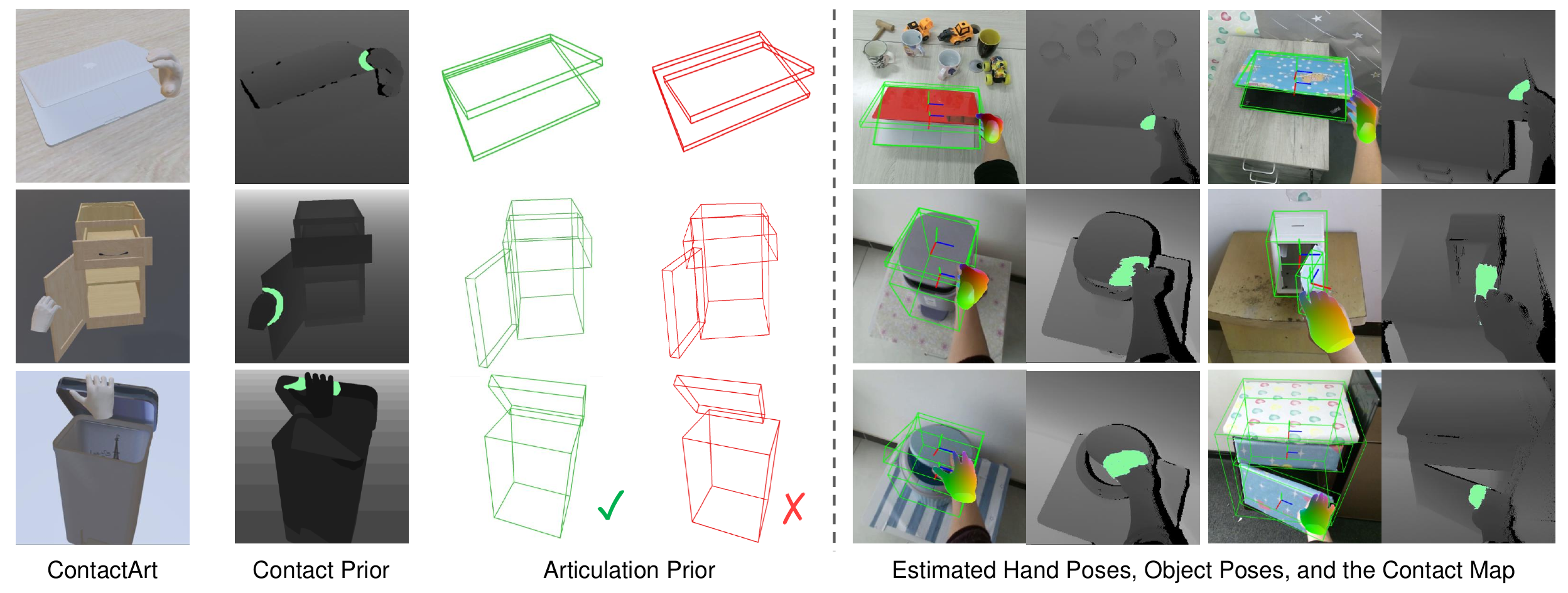}
    \vspace{-4mm}
    \captionof{figure}{\textbf{Overview.} We collect a dataset named ContactArt, which is created by human interacting with the articulated objects in a simulator, using teleoperation. Two interaction priors are learned from ContactArt: (i) a contact prior predicted by a diffusion model to improve 3D hand pose estimation; (ii) an articulation prior with a discriminator to improve category-level articulated object pose estimation. We visualize the pose estimation results in real-world data, leveraging the learned priors. } 
    \label{fig::teaser}
\end{center}

}]

\blfootnote{*: equal contributions}



\begin{abstract}
We propose a new dataset and a novel approach to learning hand-object interaction priors for hand and articulated object pose estimation. We first collect a dataset using visual teleoperation, where the human operator can directly play within a physical simulator to manipulate the articulated objects. We record the data and obtain free and accurate annotations on object poses and contact information from the simulator. Our system only requires an iPhone to record human hand motion, which can be easily scaled up and largely lower the costs of data and annotation collection. With this data, we learn 3D interaction priors including a discriminator (in a GAN) capturing the distribution of how object parts are arranged, and a diffusion model which generates the contact regions on articulated objects, guiding the hand pose estimation. Such structural and contact priors can easily transfer to real-world data with barely any domain gap. By using our data and learned priors, our method significantly improves the performance on joint hand and articulated object poses estimation over the existing state-of-the-art methods. The project is available at \url{https://zehaozhu.github.io/ContactArt/}.
\end{abstract}
\vspace{-3mm}


\section{Introduction}
The understanding of the 3D articulated structure has caught a lot of attention recently: Studies have been conducted on estimating the articulated object poses~\cite{li2019category, weng2021captra, liu2022toward}. Beyond studying the single object in isolation, understanding the interactions between human hands and articulated objects play an important role in a wide range of applications such as robotics and Augmented Reality. However, there remain several challenges for hand and category-level articulated object pose estimation given the high Degree of Freedom on poses and mutual occlusions.

Most current research focusing on articulated object pose estimation has been limited by the high cost of annotations on real-world objects~\cite{1806.06465}. To alleviate this issue, approaches to using synthetic data with cheaper annotations have been proposed~\cite{hasson2019learning,li2019category,liu2022toward,weng2021captra}. However, this inevitably introduces a sim2real gap when transferring pose estimation to images in the wild. The joint estimation of the human hand and articulated object poses makes the problem even more challenging given their mutual occlusions. Recent efforts on collecting the real-world category-level human-object 3D pose annotations~\cite{liu2022hoi4d} have largely advanced this field. However, the expensive labeling process still makes it hard to scale and it is very difficult to obtain accurate contact labels between hands and objects from observing the images. Is there a cheaper and more scalable way to obtain the hand-object interaction annotations? 

Our answer is affirmative and our key insight is that, while there is a large sim2real appearance gap, the geometric contacts between hand and objects are actually consistent across simulation and the real world~\cite{qin2022one}. In this paper, we collect the hand-object interaction data and accurate annotations by asking humans to directly play within a physical simulator using visual teleoperation (Fig.~\ref{fig::teaser} 1st column). We name this dataset \textbf{ContactArt}: \textbf{Contact} with \textbf{Art}iculation. Specifically, we deploy an off-the-shelf visual teleoperation system~\cite{qin2022one} for data collection. It only requires a single camera on a mobile device to record the human hand, which is  mapped to a MANO hand~\cite{MANO:SIGGRAPHASIA:2017}, to manipulate the articulated objects in a physical simulator. Within each object category, we collect interaction data across diverse articulated object instances. We can obtain the accurate hand-object poses and their contact points for free by reading from the simulator. This largely reduces the labeling cost from previous approaches~\cite{chao2021dexycb,liu2022hoi4d}. 

The ContactArt dataset enables us to train real-world pose estimators with free annotations. 
To minimize the sim2real gap, we learn 3D interaction priors from \mbox{ContactArt} and use them to improve the real-world hand and object pose estimation.
We train a generalizable model for each object category and evaluate the model on unseen instances. We 
propose to learn two types of 3D hand-object interaction priors, which capture how object parts are generally articulated and where humans generally touch the object for manipulation. The first prior is to learn the discriminator network, modeling the joint distribution of object part arrangement inside each object category (Fig.~\ref{fig::teaser} 3rd column). Following a GAN framework~\cite{https://doi.org/10.48550/arxiv.1406.2661}, we consider the pose estimators as the generators, and we train the discriminator by using the estimated hand and object poses as the fake data, and the ground-truth poses as the real data. The discriminator then learns how object parts ``naturally'' connect together, and we use this discriminator via back-prop to optimize the estimated object pose. The second prior is to learn a contact map diffusion generator~\cite{pmlr-v37-sohl-dickstein15} for modeling where the hand can touch the object (Fig.~\ref{fig::teaser} 2nd column). Given the input articulated object, this model predicts the plausible regions where the human hand operates (object affordance regions), using a diffusion process. With an initial hand pose estimation, we optimize the hand pose to match the estimated hand-object contact information. The two priors are complementary to each other and are used jointly to optimize both  hand and articulated object poses.

We perform our experiments on three in-the-wild articulated object datasets, HOI4D~\cite{liu2022hoi4d}, BMVC~\cite{BMVC2015_181} and RBO~\cite{1806.06465} including five categories in total. We find that with our ContactArt dataset and the proposed articulation and the contact prior, we can not only achieve large improvements over previous state-of-the-art methods of estimating articulated object poses, but also observe significant improvements in the hand pose estimation. Further, we find training on ContactArt first as a warm start then finetuning on HOI4D can bring better performance while requiring less data compared with training from scratch on HOI4D.

Our contributions include: (i) A new dataset with contact-rich hand-articulated object interaction; (ii) A contact diffusion model used to estimate the contact map of interaction; (iii) An articulation discriminator which learns articulation prior and boosts articulated object pose estimation; (iv) Substantial performance improvement on the articulated object and hand pose estimation.

\section{Related Work}

\textbf{Articulated Object Pose Estimation.} Beyond understanding single rigid objects, more attention has been put on articulated object modeling and pose estimation recently~\cite{BMVC2015_181,li2019category,weng2021captra,Wang_2019_CVPR,liu2022toward, a-sdf, shape2motion, zeng2021visual}. For example, Li \etal~\cite{li2019category} propose to perform  category-level articulated object pose estimation and evaluate their approach on unseen instances during training. Weng \etal~\cite{weng2021captra} adopt the ANCSH~\cite{Wang_2019_CVPR} to handle category-level pose tracking for both rigid and articulated objects by leveraging the RotationNet and CoordinateNet. Liu \etal~\cite{liu2022toward} reform the articulated object pose estimation setting for real-world environments and build an articulated object dataset ReArt-48. However, these approaches mainly focus on modeling the articulated object itself without considering how hands and objects contact and interact. In this paper, we study the joint pose estimation problem with hand and object together using different priors learned from our ContactArt dataset.

\begin{figure*}[t]
    \centering
    \vspace{-1.3mm}
    \includegraphics[width=0.93\textwidth]{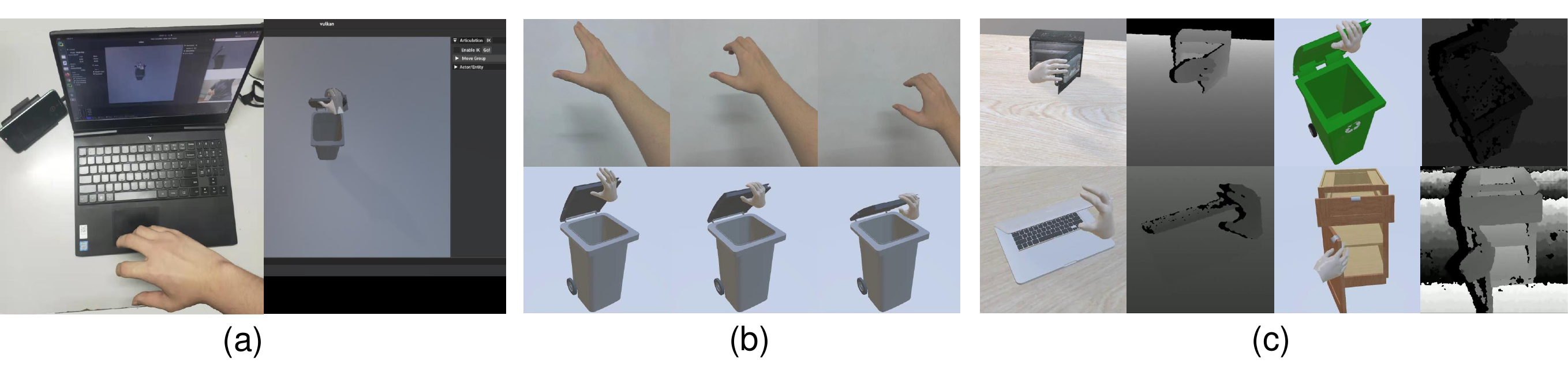}
    \vspace{-5mm}
    \caption{To \textbf{collect ContactArt}, the hardware requirement is an iPhone and a laptop. The system allows us to easily scale up the dataset without human annotation effort. We can collect manipulation sequences and render images from different camera views.}
    \label{fig::dataset}
    \vspace{-0.15in}
\end{figure*}

\textbf{Conditional Diffusion Probabilistic Models.} Recent progress on diffusion probabilistic models~\cite{sohl2015deep,ho2020denoising,song2020score,song2020denoising} has shown to be very effective in generating high-quality images. Inspired by these results, the conditional diffusion model has been widely applied in text-to-image generation~\cite{pmlr-v37-sohl-dickstein15, https://doi.org/10.48550/arxiv.2105.05233, https://doi.org/10.48550/arxiv.2102.09672, 
song2021scorebased, rombach2021highresolution, https://doi.org/10.48550/arxiv.2205.11487}, image super resolution~\cite{https://doi.org/10.48550/arxiv.2104.07636, kawar2022denoising, rombach2021highresolution} and image-to-image translation tasks~\cite{https://doi.org/10.48550/arxiv.2112.05146, https://doi.org/10.48550/arxiv.2111.05826, zhao2022egsde, baranchuk2021labelefficient}. Different from the above diffusion models which are conditioned on input image or prompt, our proposed contact diffusion model is conditioned on the point-wise feature from the point cloud.
There have been several work~\cite{baranchuk2021labelefficient, https://doi.org/10.48550/arxiv.2112.00390, wolleb2021diffusion} performing semantic segmentation with a conditional diffusion model. 
For example, Wolleb \etal~\cite{wolleb2021diffusion} uses the stochastic sampling process to implicitly group the segmentation masks of medical images. Inspired by these works, our contact diffusion model adopts the diffusion process to predict the contact map indicating where the hand should touch the articulated object.

\textbf{Adversarial Learning for Priors} While adversarial learning is initially proposed for image generation~\cite{goodfellow2020generative}, the discriminator trained with adversarial learning is also utilized in multiple tasks such as 3D human pose estimation~\cite{kanazawa2018end, kocabas2020vibe, wang2021multi, DBLP:conf/cvpr/BarsoumKL18, DBLP:conf/eccv/GuiWLM18, Aksan_2019_ICCV} and 2D human trajectory prediction~\cite{gupta2018social, kosaraju2019social, sadeghian2019sophie, amirian2019social}. Our articulation prior is inspired by~\cite{kanazawa2018end, kocabas2020vibe}, which are focusing on a specific articulation category: human. These approaches try to jointly learn a prior for what is a natural pose for humans, and how each articulated part is combined with the others. Similarly, in articulated objects, we have the upper drawer and the lower drawer are always parallel for example. Thus we propose to utilize the discriminator from adversarial learning to capture the articulation priors. 

\textbf{Hand-Object Interaction.} 
Estimating hand-object interaction has been a long-standing problem in  computer vision~\cite{hamer2010object,oikonomidis2011full,ballan2012motion,panteleris20153d,sridhar2016real}. More recently, 
a line of studies~\cite{hasson2019learning, kokic2019learning, oberweger2019generalized, tekin2019h+, zhang2019interactionfusion, hampali2020honnotate, doosti2020hope, chen2021joint, cao2021reconstructing, hasson2020leveraging, zhang2021single} use data-driven  methods to jointly estimate or reconstruct the hand and object. Another line of research studies synthesizing plausible hand-object interactions ~\cite{corona2020ganhand, karunratanakul2020grasping, jiang2021hand, brahmbhatt2019contactgrasp, grady2021contactopt, zhang2021manipnet, zhu2021toward, christen2022d}.  
Jiang et al.~\cite{jiang2021hand} propose to generate the hand grasp pose and contact map at the same time and optimize the consistency between the hand and object during test time. Our method is also related to the approach proposed by Grady et al.~\cite{grady2021contactopt} which estimates contact given a complete mesh. In contrast, our estimation is based on partial point clouds from the depth sensor which is much more challenging.

The success of these recent works is inseparable from the hand-object interaction datasets~\cite{hasson2019learning, hampali2020honnotate, brahmbhatt2019contactdb, brahmbhatt2020contactpose, taheri2020grab, chao2021dexycb, qin2022dexmv, liu2022hoi4d, fan2022articulated, grauman2022ego4d, yang2022oakink, xiang2020sapien}, which are playing crucial roles in both estimation, synthesis and robot manipulation tasks. For example, DexYCB~\cite{chao2021dexycb} and HOI4D~\cite{liu2022hoi4d} are two recent hand-object pose datasets annotated by humans, which are much more expensive compared to 2D labels. ContactDB~\cite{brahmbhatt2019contactdb} is proposed to capture the hand-object contact map with a thermal camera. But this is difficult to scale given the equipment requirement. To remove the constraints from annotation cost and hardware setup, we propose to collect the human and articulated object interaction using visual teleoperation in a physical simulator. We provide a scalable solution with free annotations from the simulator. Such geometric priors and contacts are transferable to the real world.

Vision-based manipulation teleoperation~\cite{kofman2007robot, du2012markerless, du2010robot, li2019vision, antotsiou2018task, sivakumar2022robotic, qin2022one, handa2020dexpilot} is a commonly applied technique in robotics. To reduce the device cost for scalable collection, we build our system upon~\cite{qin2022one}, using only a single camera to record the human hand to manipulate the articulated objects inside the Sapien~\cite{xiang2020sapien} simulator. While using visual teleoperation itself is not new, applying it to record hand-object poses and contacts for learning 3D priors has not been done before. We believe this can provide a new perspective on how interaction dataset can be collected for computer vision tasks.

\section{ContactArt Dataset}

To the best of our knowledge, there is only one large-scale dataset~\cite{liu2022hoi4d} including 3D hand-articulated object interaction. However, it still holds the following limitations. (i) HOI4D dataset is only captured in the egocentric view and can not generalize to the third view. (ii) The annotation of the HOI4D dataset is not accurate enough to provide contact information. Therefore we build a large hand-articulated object interaction dataset with no annotation effort and more accurate pose and contact information. 

We deploy the teleoperation system introduced in~\cite{qin2022one} to manipulate articulated objects in the Sapien~\cite{xiang2020sapien} simulation. This simulation system allows us to get accurate pose annotation and contact information using only a mobile phone and a laptop (Fig.~\ref{fig::dataset} (a)). 
Instead of using professional equipments like Kinect or Real-Sense cameras, the utilization of a mobile phone makes the data collection more accessible and easier to scale up. 
Since the teleoperation is in the simulation, the annotations can be automatically recorded, which will make it easy to scale up the size of the dataset. We train our models with this collected dataset.  While the use of the teleoperation system itself is not new, \textbf{its application in collecting 3D interaction prior data for pose estimation is novel}, compared to previous literature on collecting human demonstrations.

\textbf{Dataset collection.} 
We use the front camera of a mobile phone to stream the RGB-D video at 15 fps. The setup and collection interface are shown in Fig.~\ref{fig::dataset} (a). We adopt the teleportation system~\cite{qin2022one} which allows one to control the customized robot hand with the captured hand motion as control signals in the simulation. One can easily manipulate the objects, such as opening the drawer.
In the teleportation system, the hand is the only instance in the foreground without occlusion, making hand capture easy.
We provide a sequence of ContactArt collection process in Fig.~\ref{fig::dataset} (b). We render the RGB, depth, and segmentation images respectively, and give examples in Fig.~\ref{fig::dataset} (c). We record the object pose, bounding boxes, hand poses, and the hand-object contact regions. Note that for rendering the depth image, we apply the active stereovision depth sensor simulation proposed in~\cite{zhang2022close}, which renders realistic depth images close to the depth camera captured in the real world.

One challenge for dataset collection is that hands may touch the unexpected part. For instance, when attempting to pull the lower drawer, the hands may inadvertently move the upper drawer. We identified three types of failure cases: accidental touch, hand movement outside the boundary, and failed hand initialization. We  review each collected sequence and re-collect data if there are failure cases.

\textbf{Dataset statistics.} 
We select five common articulated object categories in our daily life including laptops, drawers, safes, microwaves and trashcans, 80 instances in total. All the object models are from Partnet dataset~\cite{https://doi.org/10.48550/arxiv.1812.02713} so it is convenient to scale up. Tab.~\ref{tab::statistics} summarizes the statistics of ContactArt comparing previous datasets. ContactArt can provide accurate annotation, rich hand-object interaction, and contact information. One can also easily render more frames by using different camera views. 

The current development on articulated object pose estimation is still in its initial stage. While the current largest HOI4D dataset~\cite{liu2022hoi4d} has 16 object categories, 8 are articulated categories, and 1 category is evaluated in the HOI4D paper.  We thoroughly evaluate 5 categories in our paper, and provide a much lower-cost manner to collect data and its accurate ground-truths. This data collection manner allows diverse object shapes, configurations, and viewpoints for rendering. We will show in the experiment that using ContactArt achieves better transfer results than HOI4D. While we agree on the importance of large-scale real-world data, we emphasize the large potential of creating diverse data cheaply and efficiently with ContactArt. 
Further, we believe our data is not the sole contribution, and we have already shown its effectiveness with \textbf{our novel approaches}.

\begin{table}
    \tablestyle{2pt}{1.1}
    \footnotesize
\begin{tabular}{ l | c  c  c  c  c}
	Dataset & HOI & Hand GT  & Multi Views & Contact Label & Frames \\
	\hline
	BMVC~\cite{BMVC2015_181} & \textcolor[rgb]{0,0.8,0}{\ding{51}}  & \textcolor{red}{\ding{53}}  & \textcolor{red}{\ding{53}} &\textcolor{red}{\ding{53}} & 8K \\
	RBO~\cite{1806.06465} & \textcolor[rgb]{0,0.8,0}{\ding{51}} & \textcolor{red}{\ding{53}}  & \textcolor{red}{\ding{53}} & \textcolor{red}{\ding{53}} & 12K\\
	ReArtMix~\cite{liu2022toward} & \textcolor{red}{\ding{53}} & \textcolor{red}{\ding{53}}  & \textcolor[rgb]{0,0.8,0}{\ding{51}} & \textcolor{red}{\ding{53}} & 100K\\
	HOI4D~\cite{liu2022hoi4d} & \textcolor[rgb]{0,0.8,0}{\ding{51}} & \textcolor[rgb]{0,0.8,0}{\ding{51}}  & \textcolor{red}{\ding{53}} & \textcolor{red}{\ding{53}} &1.44M\\
	ContactArt & \textcolor[rgb]{0,0.8,0}{\ding{51}} & \textcolor[rgb]{0,0.8,0}{\ding{51}}  & \textcolor[rgb]{0,0.8,0}{\ding{51}} & \textcolor[rgb]{0,0.8,0}{\ding{51}} &552K\\
\end{tabular}
  \vspace{-3.5mm}
  \caption{Comparison with other articulated object datasets. HOI refers to hand-object interaction and Hand GT refers to the annotation of ground truth hand pose. ContactArt allows rendering in multi-view and has accurate contact information. Statistics are performed on articulated objects. }
  \label{tab::statistics}
  \vspace{-0.13in}
\end{table}

\section{Method}
\label{sec::method}
\begin{figure*}[t]
    \centering
    \vspace{-2mm}
    \includegraphics[width=0.97\textwidth]{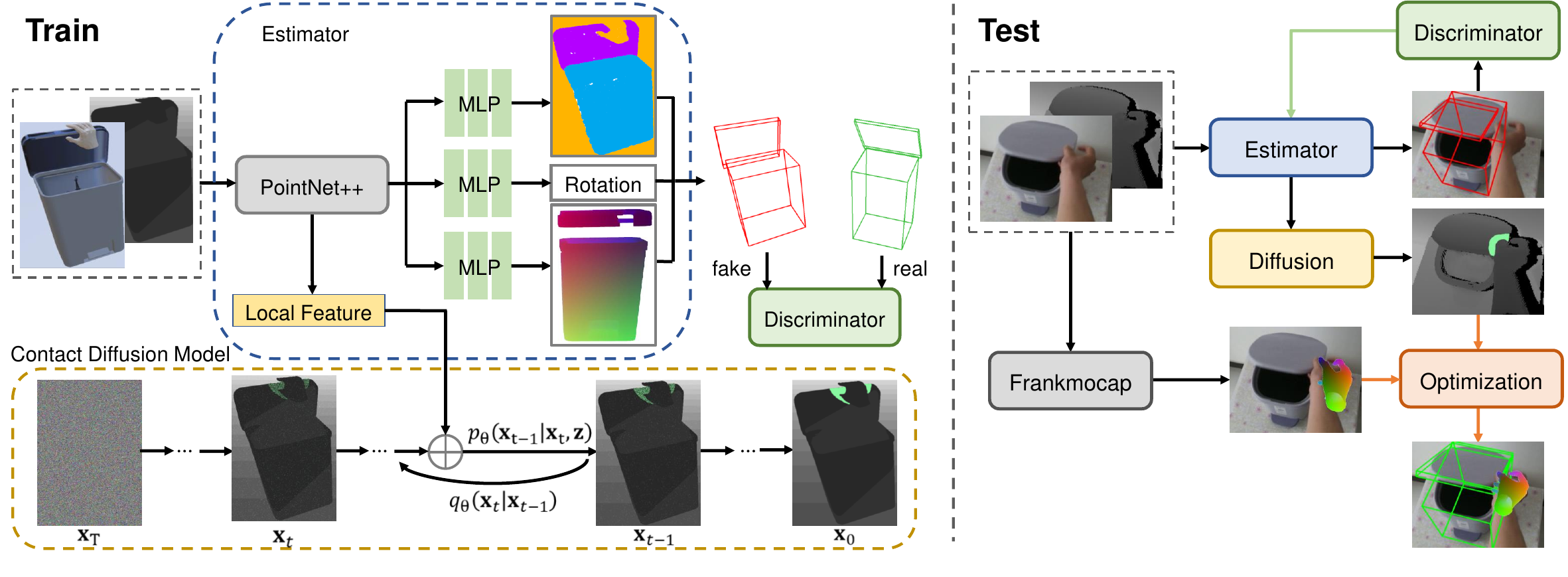}
    \vspace{-0.22in} 
    \caption{\textbf{Training and Testing framework.} \textbf{Left}: During training, we first extract a point-wise feature and regress the part segmentation, NOCS map, and part-level rotation. We then compute the per-part 3D bounding box and feed it to a discriminator. We utilize a diffusion model conditioned on the point-wise feature to estimate the contact map. We visualize the contact points in green.  $\bigoplus$ denotes concatenation. \textbf{Right}: At test time, we utilize the discriminator with fixed parameters to calculate adversarial loss and backpropagate the gradients to update the object estimator. Then we optimize hand pose by "pulling" hand closer to the contact points at the predicted contact map. }
    \label{fig::methods}
    \vspace{-0.18in}
\end{figure*}

In this paper, we target at the problem of hand and articulated object pose estimation from known categories with interaction. Compared to ~\cite{weng2021captra, liu2022toward, liu2022toward, li2019category} which only focus on the pose estimation of articulated object, our method pays attention to both hand and articulated object since they influence a lot each other during interaction. Our method takes a RGB-D image as input and output part-level 6D object pose (rotation and translation) and the hand pose parameterized by the MANO model~\cite{MANO:SIGGRAPHASIA:2017}.

We propose a NOCS-based~\cite{Wang_2019_CVPR} category-level pose estimator for articulated objects, together with two 3D interaction priors. In our framework, we train the articulated object pose estimator using both the reconstruction loss, and adversarial training with a discriminator. This \textbf{Articulation Discriminator} will serve as a prior of how object parts should be arranged together within a category. During test time, given an initial estimation from the pose estimator, we can use the discriminator to provide the gradients through back-propagation to optimize the pose of each object part. Meanwhile, to model the hand-object interaction, we also propose a diffusion-based contact map generator, which estimates the regions where the hand will touch the object, namely \textbf{Contact Diffusion Model}. We will use it as an optimization constraint to encourage the hand to reach the generated contact region. The architecture of our model is shown in Fig.~\ref{fig::methods}.

\subsection{Object Pose Estimator}
We design a multi-branch pose estimator $\mathcal{E}$ to predict the articulated object pose (Fig.~\ref{fig::methods} dotted blue box).  We first detect the hand and object and get the 2D bounding box of them with off-the-shelf method~\cite{Detectron2018} and back project the patch to point cloud $v \in \mathbb{R}^{N \times 3}$. Then we utilize PointNet++~\cite{qi2017pointnet++} to extract the point features, which include both object-aware and hand-aware information. Building on the points features, we use three separate MLPs to predict (i) the part segmentation, (ii) part-level NOCS map~\cite{Wang_2019_CVPR,li2019category}, and (iii) rotation of each part. We adopt the 6D continuous rotation representation~\cite{6drotation} for rotation. The part-level NOCS map is defined as a 3D space contained within a unit cube of each articulated part and consistently aligns to a category-level canonical orientation. 

In a forward pass, given the predictions of per-part rotation and dense correspondence between the NOCS map and point cloud, we can analytically compute translation and scale via the Umeyama algorithm. Then we leverage the computed pose (rotation, translation, scale) to transform the canonical 3D bounding box into the camera space. Different from~\cite{liu2022toward, li2019category}, by predicting the rotation of each part using a neural network, we make the prediction of pose and bounding box fully differentiable, allowing end-to-end training. This also provides opportunities for optimization using the discriminator from adversarial learning. 

Specifically, we use the cross-entropy loss ($CE$) for part segmentation. For rotation loss, we calculate the L2 distance between our prediction and ground truth in this 6D space in the form of continuous rotation representation. We use L2 distance for NOCS map loss. The object pose estimation loss can be written as,
\setlength\abovedisplayskip{0.06cm}
\setlength\belowdisplayskip{0.06cm}
\vspace{-1mm}
\begin{align}
    L_{pose}= &\lambda_{seg}\sum_i^N{CE(s_i, s_i^*)} + \lambda_{rot} \Vert r - r^* \Vert_2 \nonumber \\
    & + \lambda_{nocs}\sum_i^N{\mathds{1} (s_i^*>0)\Vert n_i - n_i^* \Vert_2} 
\end{align}
where $N$ is the number of sampled points, $s^*, s$ are the ground truth and predicted part segmentation, $n^*, n$ are the ground truth and predicted part-level NOCS maps and $r^*, r$ are the ground truth and predicted rotation. $\lambda_{nocs}$, $\lambda_{seg}$ and $ \lambda_{rot}$  are hyperparameters balancing the weights. In addition to $L_{pose}$, we introduce an adversarial loss with the Articulation Discriminator as described in the following section. 

\vspace{1.5mm}
\subsection{Articulation Discriminator}
\label{sec::discriminator}
Our model jointly learns an Articulation Discriminator $\mathcal{D}$ (Fig.~\ref{fig::methods} green box) as the articulation structure prior during training the estimator $\mathcal{E}$. This discriminator will improve the naturalness on how parts are arranged together. The discriminator takes inputs as the 3D bounding boxes, which 
are represented by 8 vertices for each part and
fully reflect the part placement rules. Furthermore, there is only a small sim2real domain gap in the 3D bounding box space.  We can calculate each part's bounding box $\hat{b}$ with the outputs from the estimator. During training, we use the estimated boxes $\hat{b} \sim p_\mathcal{E}$ as negative samples. We use the accurate bounding boxes $b$ from simulation data $p_{S}$ as positive samples. We define the loss function for the discriminator as,
\begin{equation}
    L_{\mathcal{D}}=\mathbb{E}_{b \sim p_{S}} [(\mathcal{D}(b)-1)^2]+\mathbb{E}_{\hat{b} \sim p_{\mathcal{E}}}[(\mathcal{D}(\hat{b}))^2].
\end{equation}
And the adversarial loss term for the estimator is, 
\begin{equation}
    L_{adv}=\mathbb{E}_{\hat{b} \sim p_{\mathcal{E}}}[(\mathcal{D}(\hat{b})-1)^2].
\end{equation}

\subsection{Contact Diffusion Model}
\label{sec::diffusion}
Diffusion models have shown state-of-the-art performance in generation tasks. In our work, we extend it to generate a realistic 3D contact map between the object and hand point clouds (Fig.~\ref{fig::methods} bottom). Once we get such contact information, we use it to guide the optimization of the 3D hand.

We first define the contact map. For an input point cloud set $v \in \mathbb{R}^{N \times 3}$, the contact map $x \in \mathbb{R}^{N \times 1}$ is defined as a binary vector indicates whether each point belongs to the contact region or not. We calculate the L2 distance between the points from the object and its nearest points from the hand. If this distance is smaller than a threshold, we take this point as contacted.  

We formulate the details of our contact diffusion model as follows. Let $X_0 = (x_0, z)$ denote the input, where $x_0 \in \mathbb{R}^{N \times 1}$ is the contact map and $z \in \mathbb{R}^{N \times 3}$ is the PointNet++ local feature. $T$ is the number of steps in the diffusion model and the intermediate results can be denoted as $X_t = (x_t, z)$, where $0 \leq t \leq T$. Diffusion models are composed of forward and backward processes. The forward process gradually injects random noise into the distribution, while the generative process learns to remove noise to obtain realistic samples by mimicking the reverse process. The forward process converts the original contact map distribution into a noise distribution, which can be described by the formulation,
\vspace{-1mm}
\begin{equation}
    q(x_{1:T}|x_{0}) = q(x_0) \prod \limits_{t=1}^Tq(x_t|x_{t-1}).
\end{equation}

The reverse process $p_{\theta}$ is learned by the model parameters $\theta$. Different from the forward process which simply adds noise to the contact map, the reverse process recovers the desired contact map from the input noise, encoded by the PointNet++ feature $z$. The reverse diffusion process is,
\vspace{-0.06in}
\begin{align}
p_{\theta}(x_{0:T}|z) = p(x_{T}) \prod \limits_{t=1}^Tp_{\theta}(x_{t-1}|x_t, z).
\end{align}

A parameterization trick~\cite{ho2020denoising} is used to simplify the training objective. The simplified training objective becomes,
\vspace{-0.15in}
\begin{align}
E_{X_0 \sim q(x_0), x_{1:T} \sim q(x_{1:T}|x_0, z_0) }[\sum_{t=1}^T\log{p_{\theta}(x_{t-1}|x_t, z)}].
\end{align}
Since posterior $q(x_{t-1}|x_t, x_0, z)$ is known and its
derivation is similar to the unconditional generative model, we define the $ L_{diff}$ as,
\begin{align}
    L_{diff} = \Vert \epsilon - \epsilon_{\theta}(x_t, z, t) \Vert^2.
\end{align}
Please note that the whole diffusion model is trained with the estimator in an end-to-end fashion, which takes the PointNet++ feature predicted by the estimator as input. Specifically, in our diffusion model, $\epsilon_{\theta}$ is implemented in MLP. We first concatenate PointNet++ feature $z$,  contact map $x_t$, and the time embedding in the current step t. After that, we pass the concatenated feature into MLP. Then we use the predicted noise to compute the affordance map in the next step. Same to ~\cite{https://doi.org/10.48550/arxiv.2112.00390}, we also employ multiple generations to boost performance. Finally, the total loss of the whole pipeline can be written as,
\begin{align}
L= L_{pose} + \lambda_{adv} L_{adv} + \lambda_{diff} L_{diff} ,
\end{align} 
where  $\lambda_{diff}$ and $\lambda_{adv}$ are hyperparameters balancing the diffusion model loss and the adversarial training loss.

\subsection{Test Time Adaptation}
Once we learn the \textbf{Articulation Discriminator} and the \textbf{Contact Diffusion Model}, we can improve the initial object and hand pose estimation with test time adaption. For optimizing object pose during test time, we fix the parameters of the discriminator $\mathcal{D}$ and use it to calculate the adversarial loss and backpropagate the gradients to object pose estimator $\mathcal{E}$ to boost object pose estimation. For optimizing hand pose, we employ the contact diffusion model to estimate the contact region and obtain the contact point set $C \in \mathbb{R}^{K \times 3} $ where $K$ is the number of contact points. We then optimize the MANO~\cite{MANO:SIGGRAPHASIA:2017} parameters of the hand which is initialized by the FrankMocap~\cite{rong2020frankmocap} hand pose estimator. Specifically, we minimize the chamfer distance between the hand vertices $V \in \mathbb{R}^{N\times 3}$ where $N$ is the number of vertices and the contact points,
\begin{equation}
    L_{CD}=\frac{1}{N}\sum_{v \in V} \min_{c \in C}\lVert v-c {\rVert}_2+\frac{1}{K}\sum_{c \in C} \min_{v \in V}\lVert c-v{\rVert}_2.
\end{equation}

\vspace{-0.08in}
\section{Experiments}
\subsection{Datasets}
\vspace{-0.02in}
We train our model on ContactArt and test on HOI4D~\cite{liu2022hoi4d}, RBO~\cite{1806.06465}, BMVC~\cite{BMVC2015_181} respectively. 
\textbf{HOI4D}~\cite{liu2022hoi4d} is a large-scale
hand-object interaction dataset where we can evaluate both object and hand pose estimation. We evaluate on 4 categories: safe, trashcan, laptop, and drawer.  \textbf{RBO}~\cite{1806.06465} is a collection of RGB-D video sequences. There is no annotation of hand pose so we only estimate object pose on RBO. We utilize 3 categories: laptop, microwave and drawer. \textbf{BMVC}~\cite{BMVC2015_181} includes video sequences recording articulated object with a moving camera. We evaluate object pose estimation on laptop sequence following ~\cite{weng2021captra}.

\vspace{-0.02in}
\subsection{Metrics and Methods for Comparison}
\vspace{-0.02in}
\textbf{Metrics for comparison.} For object pose estimation, we evaluate the following metrics:  5$^\circ$5cm: percentage of results with rotation and translation error smaller than 5$^\circ$ and 5cm, mIoU: the average 3D intersection over the union of ground-truth and predicted bounding boxes,  $R_{err}$: rotation error in degrees, $T_{err}$: translation error in centimeters. 
For hand pose estimation, we report mean per vertex
position error (MPVPE) and mean per joint position error (MPJPE).

\noindent \textbf{Methods for comparison.} We compare our method with two state-of-the-art image-based pose estimation works ANCSH~\cite{li2019category} and ReArtNocs~\cite{liu2022toward}, and a tracking method CAPTRA~\cite{weng2021captra}. All the baseline methods are trained on ContactArt. We also compare first training on ContactArt then finetuning on HOI4D (named Finetune) with training on HOI4D from scratch (named HOI4D*).

\begin{table*}
    \centering
    \begin{minipage}{0.54\linewidth}
    \vspace{-1.5mm}
    \tablestyle{3pt}{0.95}
    \footnotesize
    \setlength{\tabcolsep}{1.5pt}    
    \begin{tabular}{cc|cccc|cc}    
    \multicolumn{1}{c}{Category} &Metric &ANCSH&ReArtNocs&CAPTRA&Ours&HOI4D*&Finetune\\
    \shline
    \multirow{4}*{Laptop}&5$^\circ$5cm $\uparrow$   &10.54&10.60&16.35&\textbf{18.65}&61.95&\textbf{62.50}\\
    &mIoU$\uparrow$&47.8&49.52&51.57&\textbf{52.45}&65.21&\textbf{66.45}\\
    &$R_{err}$ $\downarrow$ &24.06&23.40&18.52&\textbf{17.73}&6.24&\textbf{5.20}\\
    &$T_{err}$ $\downarrow$&23.35&22.41&19.75&\textbf{18.91}&7.68&\textbf{7.17}\\
    \hline
    \multirow{4}*{Trashcan}&5$^\circ$5cm$\uparrow$&0&0&\textbf{3.05}&2.70&22.8&\textbf{24.2}\\
    &mIoU$\uparrow$&38.38&39.30&41.50&\textbf{41.95}&63.65&\textbf{64.95}\\
    &$R_{err}$ $\downarrow$&26.93&25.57&21.97&\textbf{21.43}&7.05&\textbf{5.98}\\
    &$T_{err}$ $\downarrow$&36.72&36.65&\textbf{30.70}&30.75&7.75&\textbf{7.22}\\
    \hline
    \multirow{4}*{Safe}&5$^\circ$5cm$\uparrow$&1.66&5.30&4.65&\textbf{8.43}&32.05&\textbf{33.40}\\
    &mIoU$\uparrow$&46.20&47.05&46.83&\textbf{47.96}&62.31&\textbf{63.50}\\
    &$R_{err}$ $\downarrow$&18.63&16.91&16.43&\textbf{16.24}&5.74&\textbf{5.11}\\
    &$T_{err}$ $\downarrow$&17.52&16.51&16.55&\textbf{15.66}&6.85&\textbf{6.50}\\
    \hline
    \multirow{4}*{Cabinet}&5$^\circ$5cm$\uparrow$&0.50&0&0.16&\textbf{1.00}&22.07&\textbf{23.87}\\
    &mIoU$\uparrow$&49.83&49.90&49.76&\textbf{50.40}&64.43&\textbf{66.71}\\
    &$R_{err}$ $\downarrow$ &18.94&19.52&19.87&\textbf{17.84}&6.46&\textbf{5.75}\\
    &$T_{err}\downarrow$&23.33&23.28&24.23&\textbf{22.60}&6.03&\textbf{5.83}\\
    \shline
    \multirow{4}*{Average}&5$^\circ$5cm$\uparrow$&3.17&3.97&6.05&\textbf{7.69}&34.72&\textbf{35.99}\\
    &mIoU$\uparrow$&45.55&46.44&47.41&\textbf{48.19}&63.90&\textbf{65.40}\\
    &$R_{err}$ $\downarrow$&22.14&21.35&19.20&\textbf{18.31}&6.37&\textbf{5.51}\\
    &$T_{err}$ $\downarrow$&25.23&24.71&22.81&\textbf{21.98}&7.08&\textbf{6.68}\\
    \end{tabular}
    \vspace{-0.15in}
    \caption{Quantitative comparison of  object pose estimation on HOI4D. Our method and dataset could both improve the estimation performance.}
    \label{tab:hoi4d}
    \end{minipage}
    \
    \begin{minipage}{0.45\linewidth}
    \vspace{-4.1mm}
    \tablestyle{5pt}{1.0}
    \footnotesize
    \setlength{\tabcolsep}{3pt}        
    \begin{tabular}{cc|cccccc}
    \multicolumn{1}{c}{ Category} &Metric&Ansch&ReArtNocs&CAPTRA&Ours\\
    \shline
    &5$^\circ$5cm$\uparrow$&1.45&0.75&4.02&\textbf{4.70}\\
    BMVC&mIoU$\uparrow$&54.32&54.93&60.25&\textbf{61.22}\\
    Laptop&$R_{err}$ $\downarrow$&26.72&24.04&19.08&\textbf{17.30}\\
    &$T_{err}$ $\downarrow$&18.58&18.05&12.34&\textbf{11.45}\\
    \hline
    &5$^\circ$5cm$\uparrow$&23.01&29.12&33.12&\textbf{33.83}\\
    RBO&mIoU$\uparrow$&49.85&51.11&51.77&\textbf{52.95}\\
     
    Laptop&$R_{err}$ $\downarrow$&11.39&11.56&10.89&\textbf{10.76}\\
     
    &$T_{err}$ $\downarrow$&9.60&8.95&7.12&\textbf{6.65}\\
    \hline
    &5$^\circ$5cm$\uparrow$&43.02&47.51&55.31&\textbf{57.66}\\
     
    RBO&mIoU$\uparrow$&69.21&70.65&71.45&\textbf{73.05}\\
     
    Microwave&$R_{err}$ $\downarrow$&7.28&6.56&10.89&\textbf{4.69}\\
     
    &$T_{err}$ $\downarrow$&5.20&4.87&4.91&\textbf{4.70}\\
    \hline
    &5$^\circ$5cm$\uparrow$&22.02&26.98&\textbf{33.79}&28.23\\
     
    RBO&mIoU$\uparrow$&53.93&54.83&55.01&\textbf{56.34}\\
     
    Drawer&$R_{err}$ $\downarrow$&8.27&8.34&8.01&\textbf{7.85}\\
     
    &$T_{err}$ $\downarrow$&14.93&13.45&13.10&\textbf{12.60}\\
    \end{tabular}
    \vspace{-0.16in}
    \caption{Quantitative comparison of object pose estimation on BMVC and RBO datasets. Given to the articulation prior the model learnt, our method can achieve the best performance on all categories.}
    \vspace{0.02in}
    \label{tab:bmvc}
    
    \end{minipage}
    
\end{table*}

\begin{figure*}[htp]
    \centering
    \vspace{-0.17in}
    \includegraphics[width=0.89\textwidth]{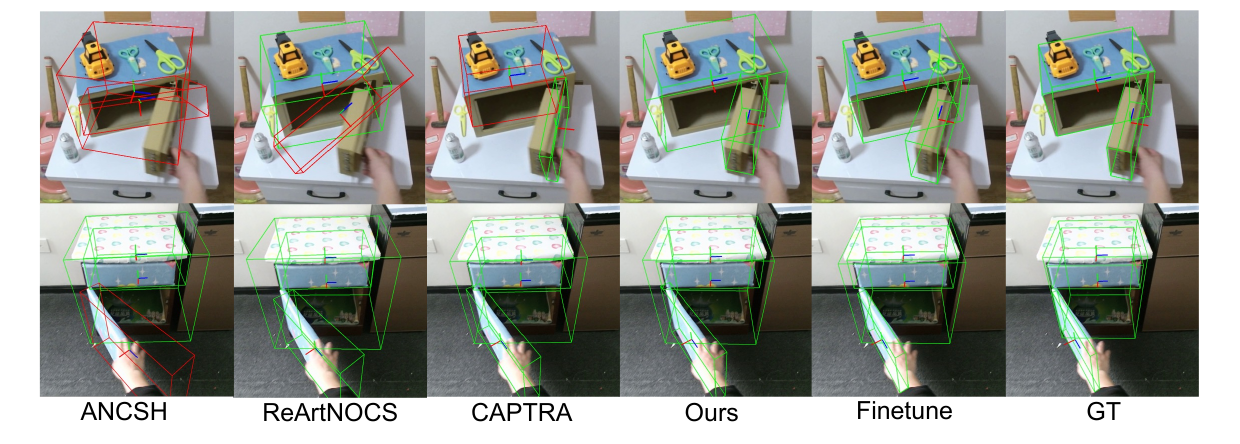}
    \vspace{-0.25in}
    \caption{Qualitative comparison of object pose estimation. We use red box to indicate error larger than 10$^\circ$ or 10 cm. Image-based baselines fail to get an accurate pose. And our method also performs better than the tracking-based method~\cite{weng2021captra}. }
    \vspace{-0.16in}
    \label{fig::compare}
\end{figure*}

\begin{figure}
    \centering 
    \vspace{-0.12in}
    \includegraphics[width=0.45\textwidth]{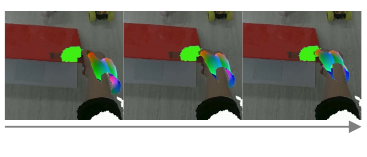}
    \vspace{-0.36in}
    \caption{Optimization process. The hand is reaching the predicted contact map and getting to the correct pose.}
    \label{fig::optim}
    \vspace{-0.15in}
\end{figure}

\begin{table}
    \tablestyle{1.5pt}{1}
    \footnotesize
    \centering
    \begin{tabular}{c|cccccc}
    Metric & Frankmocap~\cite{rong2020frankmocap} & MG~\cite{lin2021mesh} & Affine & Regression & Closest & Ours\\
    \shline
    MPVPE$\downarrow$ & 71.6 & 67.4 & 54.1 & 52.4 & 60.5 & \textbf{49.9}\\
    MPJPE$\downarrow$ & 64.3 & 65.3 & 45.9 & 44.8 & 56.1 & \textbf{41.9}\\
    \end{tabular}
    \vspace{-0.13in}
    \caption{Quantitative comparison of hand pose estimation on HOI4D. Our MLP-based contact diffusion model performs best.}
    \label{tab:hand}
    \vspace{-0.15in}
\end{table}

\begin{figure*}[!t]
    \centering 
    \vspace{-3.3mm}
    \includegraphics[width=0.87\textwidth]{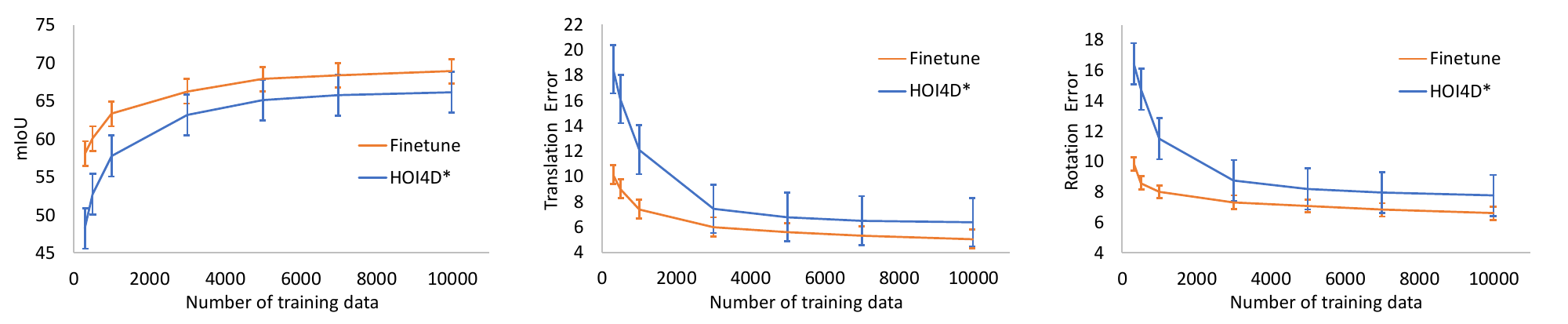}
     \vspace{-0.18in}
    \caption{Ablation on the amount of training data compared between Finetuning (Finetune) and training from scratch (HOI4D*). Finetuning can always achieve better performance with much fewer data. }
    \label{fig::curve}
     \vspace{-0.16in}
\end{figure*}

\vspace{-0.02in}
\subsection{Object Pose Estimation Comparison}
\vspace{-0.02in}
We summarize the quantitative articulated object pose estimation results on HOI4D in Tab.~\ref{tab:hoi4d}. Compared with the other methods, ours has the lowest average rotation and translation error, the highest mIoU and 5$^\circ$5cm. The last two columns shows results of training from scratch and finetuning respectively. Finetuning performs better than training from scratch for all the metrics, which demonstrated ContactArt can serve as a ``prior'' for pose estimation and could be used as a warm start for other datasets with smaller sizes.

We visualize the qualitative comparisons of articulated object pose estimation in Fig.~\ref{fig::compare}. Following~\cite{li2019category}, we utilize 10$^\circ$10cm as a threshold and use red color to indicate the results larger than this threshold and green to indicate the one within it. We observe that our method is the most accurate and robust compared with the other methods, especially on challenging layouts or camera views.

We summarize the quantitative comparisons on BMVC and RBO in Tab.~\ref{tab:bmvc}. Our method outperforms the baselines on all the categories across all the metrics. Though our model is trained on ContactArt with rich hand-object interaction but it still works well on BMVC.

\subsection{Hand Pose Estimation Comparison}
\vspace{-1mm}
For hand pose estimation,  we design an effective post-process pipeline. Specifically, we take off-the-shelf motion capturer Frankmocap~\cite{rong2020frankmocap} and our test time adaption to improve the hand estimation based on Frankmocap.  
We compare the following baselines: \textit{MG}, a state-of-the-art hand pose estimator Mesh Graphormer~\cite{lin2021mesh}; \textit{Regression}, a non-diffusion baseline where we use MLPs to decode the point feature; \textit{Affine}, where we change the MLP architecture of the diffusion model to ConcatSquash layers~\cite{grathwohl2019ffjord}; \textit{Closest}, where we simply optimize the hand towards the closest object vertices. 
The methods leveraging contact map to optimize hand (Ours, Affine, Regression) outperform methods without contact map(Frankmocap, MG, Closest). Among the three contact map estimator, our method which employs MLP-based diffusion model performs best. We also visualize the optimization process in Fig.~\ref{fig::optim}.  The contact map estimated by the diffusion model can anchors the hand to a more reasonable space. The articulation and contact prior inject a shared prior into the PointNet for mutual benefit, jointly enhancing object and hand pose estimation.


\subsection{Generalization Comparison}
We conduct two experiments to show training on ContactArt performs better for out-of-distribution data. 

\noindent \textbf{Comparison with HOI4D}.
We compare using HOI4D and ContactArt as training dataset, and report the test performance on BMVC and RBO datasets in Tab.~\ref{tab:training_set}.  The model trained on ContactArt adapts better to unseen scenes than the one trained on HOI4D significantly. This is mainly due to more diverse data with accurate label in ContactArt.
 
\noindent \textbf{Comparison with other pretrained models}.
We report the performance of the public pretrained model of CAPTRA and ANCSH on the Laptop of HOI4D in Tab.~\ref{tab:pretrained_baseline}. These two are trained on their own synthetic datasets rendered by ~\cite{xiang2020sapien}. Our method trained on ContactArt is significantly more accurate than previous SOTA. The datasets of CAPTRA and ANCSH contain only articulated objects, so they can not generalize to scenarios with rich hand-object interaction.

\begin{table}
    \tablestyle{1.5pt}{1}
    \footnotesize
    \centering
    \begin{tabular}{cc|cccccc}
    \multicolumn{1}{c}{Test set} &Training set&  5$^\circ$5cm$\uparrow$ & mIoU$\uparrow$ & $R_{err}$ $\downarrow$ & $T_{err}$ $\downarrow$\\
    \shline
    \multirow{2}*{BMVC} & Hoi4D & 0 & 37.45 & 35.77 & 39.19\\
      & ContactArt & \textbf{4.70} & \textbf{61.22} & \textbf{17.30} &\textbf{11.45}\\
    \hline
    \multirow{2}*{RBO} & Hoi4D & 5.12 & 45.33 & 25.35 &  21.94\\
     & ContactArt & \textbf{33.83} & \textbf{52.95} & \textbf{10.76} & \textbf{6.65}\\
    \end{tabular}
    \vspace{-0.17in}
    \caption{Quantitative comparison of different training sets. Training on ContactArt achieves best performance on unseen test sets.}
    \label{tab:training_set}
    \vspace{-0.04in}
\end{table}

\begin{table}
    \tablestyle{1.5pt}{1}
    \footnotesize
    \centering
    \begin{tabular}{c|cccccc}
    Method&  5$^\circ$5cm$\uparrow$ & mIoU$\uparrow$ & $R_{err}$ $\downarrow$ & $T_{err}$ $\downarrow$\\
    \shline
    Pretrained ANCSH & 6.21 & 43.12 & 35.90 & 28.13\\
    Pretrained CAPTRA & 8.57 & 41.54 & 31.11 &24.66\\
    Ours with ContactArt & \textbf{18.65} & \textbf{52.45} & \textbf{17.73} &\textbf{18.91}\\
    \end{tabular}\
    \vspace{-4.1mm}
    \caption{Quantitative comparison of our methods and baselines trained on their own dataset.}
    \label{tab:pretrained_baseline}
    \vspace{-0.01in}
\end{table}

\begin{table}
    \tablestyle{2.5pt}{0.98}
    \vspace{-0.07in}
    \footnotesize
    \centering
    \begin{tabular}{c|cc|cc|cc|cc}
    
      \multirow{2}*{Metric} & \multicolumn{2}{c|}{CAPTRA} & \multicolumn{2}{c|}{Ours} & \multicolumn{2}{c|}{HOI4D*} & \multicolumn{2}{c}{Finetune}  \\
    {}&w/o tta&w/ tta&w/o tta&w/ tta&w/o tta&w/ tta&w/o tta&w/ tta\\
    \shline
    5$^\circ$5cm $\uparrow$&16.35&\textbf{16.70}&18.00&\textbf{18.65}&62.05&\textbf{62.50}&61.55&\textbf{61.95}\\
     
    mIoU$\uparrow$&51.50&\textbf{51.65}&52.15&\textbf{52.45}&66.25&\textbf{66.45}&\textbf{65.25}&65.20\\
     
    $R_{err}$ $\downarrow$&\textbf{18.22}&18.3&18.26&\textbf{17.73}&5.54&\textbf{5.20}&6.42&\textbf{6.24}\\
     
    $T_{err}$ $\downarrow$&19.75&\textbf{19.60}&19.10&\textbf{18.90}&7.35&\textbf{7.20}&7.75&\textbf{7.70}\\
    \end{tabular}
    \vspace{-0.12in}
    \caption{Evaluation for test time adaption(TTA), which can benefit all methods. HOI4D* denotes training on HOI4D from scratch. }
    \label{tab:example}
    \vspace{-0.16in}
\end{table}

\vspace{-0.05in}
\subsection{Ablation Study}

\textbf{Ablation on the amount of training data}.
ContactArt could serve as a warm start before training on other datasets. To prove it, we perform ablation study on training with different amounts of data in Fig.~\ref{fig::curve}. In general, finetuning is always better than training from scratch with the same amount of data. The fewer data we give, the larger improvements our model can achieve. We can also observe that finetuning with 300, 1k, and 3k images are better than training with 1k, 3k, and 10k images respectively. We only need one-third of the data to achieve comparable performance if using ContactArt as a warm start. This is of great importance when we only have a small amount of in-the-wild data. One can first train on ContactArt and quickly adapt to new test scenarios.

\begin{figure}
    \centering
    \vspace{-0.24in}
    \includegraphics[width=0.93\columnwidth]{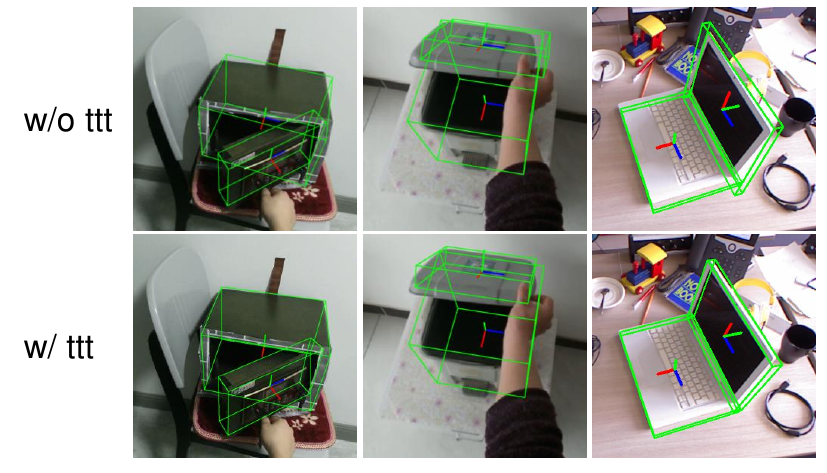}
    \vspace{-0.18in}
    \caption{Comparison between our method with and without TTA. TTA can improve the layout naturalness of each articulated part. }
    \label{fig::ttt}
    \vspace{-0.2in}
\end{figure}

\noindent \textbf{Ablation on Test time adaption (TTA).} Our discriminator is highly scalable and can be plugged into any pose estimation method, such as CAPTRA. We select four methods, CAPTRA; our method; HOI4D*; and Finetune. We add the discriminator to each method and apply the TTA to each. In Tab.~\ref{tab:example}, we compare the one with and without TTA on laptop of HOI4D. We observe that the articulation discriminator and TTA improve all the methods. The TTA
mechanism enables the estimator to adapt to various test scenes since the discriminator can learn an invariant prior of the articulation structure, specifically the layout of the bounding boxes. In Fig.~\ref{fig::ttt}, we visualize the results of our method with and without TTA. After TTA, the drawers are parallel and the laptop's screen and keyboard are at similar sizes. 

\vspace{-0.05in}
\section{Conclusion}
In this paper, we present ContactArt, an interaction- and contact-rich and easily scalable dataset. We further propose to use a discriminator as an articulation prior to improve the articulated object pose estimation. We introduce a contact diffusion model to estimate the contact map between the hand and articulated objects, which can be utilized to optimize the hand pose estimation. Extensive experiments demonstrate the effectiveness of our ContactArt dataset, the articulation prior and the contact prior. 

\noindent \textbf{Limitation.}  Our method can not generalize to novel categories since the learnt articulation priors are object-specific. Our method relies more on depth data instead of the visual textures. Our simulator presents a challenge when attempting to capture objects smaller than hands, such as scissors.

{\small
\bibliographystyle{ieee_fullname}
\bibliography{egbib}
}

\end{document}